\documentclass{article}

\usepackage{arxiv}

\usepackage[utf8]{inputenc} 
\usepackage[T1]{fontenc}    
\usepackage{hyperref}       
\usepackage{url}            
\usepackage{booktabs}       
\usepackage{amsfonts}       
\usepackage{nicefrac}       
\usepackage{microtype}      
\usepackage{lipsum}
\usepackage{graphicx}
\usepackage{subcaption}
\usepackage{amsmath}
\graphicspath{ {./images/} }

\title{Enhancing Orthopox Image Classification Using Hybrid Machine Learning and Deep Learning Models}

\author{
 Alejandro Puente-Castro \\
  Faculty of Computer Science, CITIC\\
  University of A Coruna\\
  A Coruna, Spain 15007 \\
  \texttt{a.puentec@udc.es} \\
  \And
 Enrique Fernandez-Blanco \\
  Faculty of Computer Science, CITIC\\
  University of A Coruna\\
  A Coruna, Spain 15007 \\
  \And
 Daniel Rivero \\
  Faculty of Computer Science, CITIC\\
  University of A Coruna\\
  A Coruna, Spain 15007 \\
   \And
 Andres Molares-Ulloa \\
  Faculty of Computer Science, CITIC\\
  University of A Coruna\\
  A Coruna, Spain 15007 \\
}

\begin{document}
\maketitle
\begin{abstract}
Orthopoxvirus infections must be accurately classified from medical pictures for an easy and early diagnosis and epidemic prevention. The necessity for automated and scalable solutions is highlighted by the fact that traditional diagnostic techniques can be time-consuming and require expert interpretation and there are few and biased data sets of the different types of Orthopox. In order to improve classification performance and lower computational costs, a hybrid strategy is put forth in this paper that uses Machine Learning models combined with pretrained Deep Learning models to extract deep feature representations without the need for augmented data. The findings show that this feature extraction method, when paired with other methods in the state-of-the-art, produces excellent classification outcomes while preserving training and inference efficiency. The proposed approach demonstrates strong generalization and robustness across multiple evaluation settings, offering a scalable and interpretable solution for real-world clinical deployment. 
\end{abstract}



\section{Introduction}\label{introduction}

Despite widespread vaccinations that can remove a wide range virus, several human-pathogenic orthopoxviruses remain in circulation. In recent years, new orthopoxviruses have emerged \cite{babkin2022update}, some of which, such as Monkeypox (mpox), can infect humans \cite{mitja2023monkeypox}. Since the worldwide smallpox vaccination program ended more than 20 years ago, mpox has emerged as the most common orthopoxvirus infection, with an increasing number of verified cases \cite{rahmani2024monkeypox}. For example, a characteristic symptom of Monkeypox infection is the presence of skin blisters, and the disease carries a mortality rate ranging from 0--10\%. Furthermore, a sizable section of the world's population lacks immunity to smallpox and other zoonotic orthopoxvirus illnesses due to the cessation of smallpox vaccines following its eradication in the decade of 1980. Because of this change in population immunity, orthopoxviruses have more opportunity to propagate among people, which calls into question long-held beliefs about their ecology, host range, and likelihood of reoccurring outbreaks \cite{shchelkunov2013increasing}.

Currently, the most prominent outbreaks are those of Monkeypox, which the World Health Organization has declared a global public health emergency. The main reason for this decision is that the symptoms have milder clinical presentations than before. This situation makes it more difficult to identify and interrupt transmission, specially in less developed countries \cite{nuzzo2022declaration}.

Early disease diagnosis is essential because of the above listed problems. As a result, the disease can be identified sooner to treat it and lessen its spread. The range of symptoms among all orthopoxvirus types, which may initially be similar to one another yet have unique characteristics, is a significant obstacle \cite{diaz2021disease}. In order to address these issues, experts are increasingly using artificial intelligence (AI) models \cite{brewka1996artificial}. Therefore, it is possible to build models that can generalize information about the many forms of Orthopox using these new methodologies \cite{chadaga2023application}.

To increase diagnostic precision and health response, models should take into account the range of cutaneous symptoms caused by orthopoxviruses. Lesions from these infections may appear in various phases, such as crusts, papules, or macules, which might be mistaken for those from other dermatological conditions \cite{pauli2010orthopox}. Furthermore, each patient has a different distribution and progression of lesions, needing the analysis of morphological, temporal, and distribution of the pixel values on the skin using models. As a result, the models require a large number of training instances with various phenotypic characteristics, which is expensive in terms of both time and resources \cite{zha2025data}.

In order to reflect these situations and to train AI-based models, standardized medical data sets are used. In the case of Orthopox, one of the most popular is Monkeypox Skin Images Dataset (MSID) proposed by Bala et al \cite{bala2023monkeynet}. This dataset contains 4 classes where one is the control group and the other three are skin pathologies. All classes take into account ethnicity and different angles and body parts in order to have as little bias as possible. In this way, a wide variety of phenotypic observations can be perceived to train the models.

Nevertheless, it is important to observe that there is an imbalance between classes in many datasets, such as MSID. It is challenging for the models to learn the phenotypic characteristics unique to each variant when there are classes that are underrepresented compared to the others. In the state of the art, some authors decide to create synthetic data, as covered in Section \ref{background}. There is a chance that this kind of data will have dependencies between the training and test sets, or that the new data won't be verified by professionals.

The main objective of this article is to provide a fast image categorization technique for both healthy individuals and various orthopoxvirus variants. To do this, Deep Learning (DL) models \cite{lecun2015deep} are used to extract features from the pictures, which are then used to train more straightforward and effective Machine Learning (ML) models \cite{bishop2006pattern} without the need of balanced data or synthetic data. The key highlights are:

\begin{itemize}
    \item Proposal of a preprocessing pipeline for orthopoxvirus image classification with any type of image data.
    \item Use of ResNet-based feature extraction to reduce computational cost and training time.
    \item Analysis of class imbalance impact and mitigation through SMOTEENN and data augmentation techniques.
    \item Identification of overestimation risks due to test set augmentation in prior studies.
\end{itemize}

In addition to this introductory Section \ref{introduction}, this article has the following structure: Section \ref{background} summarizes the most recent and relevant results in the state of the art; Section \ref{methods} presents the materials and methods used during the experimentation; Section \ref{results} summarizes the results obtained from the experimentation; finally, Section \ref{conclusions} summarizes the conclusions obtained from the results and possible future developments.

\section{Background}\label{background}

The integration of AI techniques into medical fields has significantly advanced diagnostic accuracy and decision-making processes. In particular, AI-based models have demonstrated promising results in dermatology, assisting in the identification of skin conditions, and enhancing early intervention strategies \cite{hogarty2020artificial}.

\subsection{Artificial Intelligence Applied to Dermatology}

Starting in 2023, Tahir et al. created a DL model for skin cancer detection in dermoscopic images \cite{tahir2023dscc_net}. By using this type of specialized imaging they can highlight skin elements. However, the cost of obtaining these images is high due to the need for equipment. Also, in their results they show an overtraining that they had to solve with Dropout layers. The same kind of imaging was employed by Ajmal et al. to identify skin cancer using DL \cite{ajmal2023bf2sknet}. They also provide an understandable approach for visualizing skin cancer-affected areas. In their results they do not seem to consider the imbalance between classes, so they present macro data on accuracy, which they then compare with other authors. This metric is very dependent on unbalance, so it is also necessary to contrast other metrics. Tembhurne et al. present a novel approach combining ML and DL techniques in a hybrid model for skin cancer detection \cite{tembhurne2023skin}. However, the study is limited by the small size of the data set used, which could affect the generalizability of the results to more diverse populations. Also, they show very high results in a voting mechanism where one ML model is trained with data augmentation and the other models are not. Because of this, it is very likely that almost all the decision will fall on that model and it is not necessary to use the others in the voting.

A DL-based method for identifying skin lesions brought on by STDs was proposed by Soe et al. in 2024 \cite{soe2024evaluation}. The outcomes of using merely damaged photographs and the outcomes of adding metadata to the images are assessed and contrasted in their system. They even simplify the problem to two classes, thus mixing different pathologies, which adds a lot of noise in the system. The Ant Colony Optimization technique was used by Sarwar et al. that same year to segment malignant skin lesions. They want to optimize a hybrid system made up of two DL models with their optimization technique \cite{sarwar2024skin}. Their results are overestimated because, as they detail in their algorithm, they first perform data augmentation and then do the data splitting. Therefore, there is data in test that is dependent on data in train. The same overestimation can be seen in the paper of Deng. In it, Deng suggested employing Transfer Learning to maximize the training offered by the DL system. As the article states, using pre-trained models may prevent developing the most accurate representations of skin lesions \cite{deng2024lsnet}.

Later in 2025, Ozdemir et al. designed a DL model with a focal mechanism to detect skin cancer \cite{ozdemir2025innovative}. Similarly, Chen et al. used DL with an attention mechanism for the diagnosis of skin damage \cite{chen2025pigmented}. In their results they not only manage to detect cancer lesions, but they are also able to segment the skin areas where these lesions are located. Also with an attention mechanism, Vuran et al. propose a DL system that makes use of Transformers to classify skin lesions of different diseases, including Monkeypox \cite{vuran2025multi}. Thus, they propose a system that prioritizes detecting diseases that may become pandemic. While the use of attention mechanisms, focal attention, and Transformer-based architectures has shown improvements in the accuracy of skin lesion classification models, their adoption introduces several critical risks. First, these approaches often significantly increase computational complexity and the number of parameter. Second, the inherent lack of interpretability of such architectures limits clinical trust, as they are rarely accompanied by clear visual explanations that allow dermatologists to understand the model's decisions.

\subsection{Artificial Intelligence Applied to Orthopox}

Over the past few years, a large number of AI-based systems specific for Orthopox detection for have emerged. In 2023 Bala et al. proposed a convolutional model called MonkeyNet along with the first standardized orthopoxvirus-based dataset \cite{bala2023monkeynet}. The main goal is to assist specialists in the early diagnosis of these diseases. However, to obtain robust results, they need to create synthetic cases, which may not be validated by experts. In the same year, Alharbi5 et al. also propose the use of Artificial Neural Networks (ANN) but from the Transfer Learning approach to optimize training and improve state-of-the-art results \cite{alharbi2023diagnosis}. However, they also need to create synthetic cases from real images. Eliwa et al. tried in 2023 to increase the number of examples, but make use of standardized methods such as SMOTEEN to reduce possible biases \cite{eliwa2023utilizing}. In addition, they make use of feature selection techniques to optimize the dimensionality of the data.

Later, in 2024, Maqsood et al. proposed a hybrid methodology that combines Transfer Learning with traditional machine learning techniques by using features extracted from pre-trained convolutional neural networks to train Support Vector Machine (SVM) models \cite{maqsood2024mox}. This strategy aims to strike a balance between model complexity and efficiency, making it suitable for clinical environments with limited computational resources. In the same year, Kundu et al. introduced a novel approach by applying Federated Learning for Monkeypox classification, allowing the training of models across distributed data sources without transferring sensitive patient information \cite{kundu2024federated}. This method significantly improves data privacy and security but also introduces limitations due to the insufficient amount of data that can be synthesized using CycleGAN, their chosen data augmentation technique. Also in 2024, Asif et al. presented an ensemble method based on multiple Artificial Neural Networks (ANNs) to enhance classification robustness and interpretability \cite{asif2024cgo}. By aggregating the outputs of several independently trained networks, their system achieves more consistent predictions; however, the effectiveness of the ensemble depends heavily on the choice of combination algorithm, which can influence both performance and reliability.

More recently, in 2025, Das et al. introduced a novel diagnostic framework that integrates Transfer Learning with an ensemble of Artificial Neural Networks (ANNs), aiming to enhance both the efficiency and accuracy of Monkeypox classification systems \cite{das2025novel}. By leveraging pre-trained models for initial feature extraction and then aggregating the predictions of multiple specialized ANNs, their approach captures a broader range of discriminative patterns from medical images, thereby improving robustness and reducing the risk of overfitting. This hybrid setup not only accelerates training by reusing learned representations but also benefits from the complementary strengths of each neural network in the ensemble. However, the authors emphasize that the success of this architecture is highly dependent on the availability of large, diverse, and representative datasets, as insufficient data variability can limit the generalizability and real-world applicability of the system across different populations and imaging conditions.

To overcome the challenge, all of the aforementioned papers typically need complex models or large amounts of varied data. The generation of data is the main issue since experts do not validate it. It is possible to create dependencies between data, which can be thought of as data dependent on tests and trains. Furthermore, as developers of extremely complicated models point out, the requirement for a large and diverse set of data may prevent their models from producing reliable conclusions. The concept presented in this study aims to create a straightforward, effective model without the requirement for data generation.

\section{Materials and Methods}\label{methods}

For the development of the solution, the elements to be taken into account to carry out the experiments have been identified. In addition, a pipeline has been defined to obtain the results.


In order to achieve the main objective it is necessary to take into account all aspects necessary for its completeness and evaluation. That is, it is necessary to retrieve data, preprocess the data, create AI models, train the models, and evaluate the models. 


\subsection{Data Retrieval}

Monkeypox Skin Images Dataset (MSID) proposed by Bala et al. in 2023 was chosen as the dataset for the experiments \cite{bala2023monkeynet}. The reason for choosing this dataset is that it presents images of different pathologies and of healthy individuals. Moreover, it is the one used by most authors in the state-of-the-art.

\begin{figure}[h!]
    \centering
    \begin{subfigure}[b]{0.4\textwidth}
        \centering
        \includegraphics[width=\textwidth]{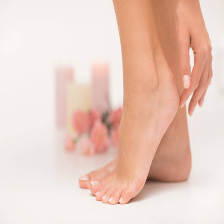}
        \caption{Normal}
    \end{subfigure}
    \hfill
    \begin{subfigure}[b]{0.4\textwidth}
        \centering
        \includegraphics[width=\textwidth]{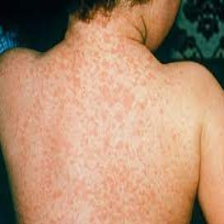}
        \caption{Measles}
    \end{subfigure}

    \begin{subfigure}[b]{0.4\textwidth}
        \centering
        \includegraphics[width=\textwidth]{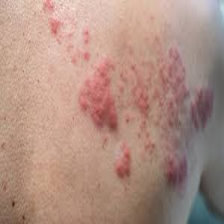}
        \caption{Chickenpox}
    \end{subfigure}
    \hfill
    \begin{subfigure}[b]{0.4\textwidth}
        \centering
        \includegraphics[width=\textwidth]{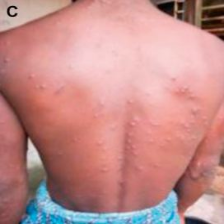}
        \caption{Monkeypox}
    \end{subfigure}
    
    \caption{Examples for each class from the original dataset \cite{bala2023monkeynet} showing the variety of ethnicities, body parts, and angles available.}
    \label{fig:data}
\end{figure}

The dataset consists of 770 images of 224$\times$224 pixels in PNG format. It presents unbalance between the 4 classes (Table \ref{tab:dataset}) it contains: normal, measles, chickenpox, monkeypox. In addition to this imbalance, it presents great phenotypic variety where different types of manifestations can be seen in people of different ethnicities and in different parts of the body. All these images are captured from different angles (Figure \ref{fig:data}).

\begin{table}[h!]
    \centering
    \begin{tabular}{|l|l|}
    \hline
        \textbf{Class} & \textbf{Quantity} \\ \hline
        Normal         & 293 images        \\ \hline
        Measles        & 91 images         \\ \hline
        Chickenpox     & 107 images        \\ \hline
        Monkeypox      & 279 images        \\ \hline
    \end{tabular}
    \caption{Summary table with the number of images per class.}
    \label{tab:dataset}
\end{table}

\subsection{Data Preprocessing}

Given the class imbalance and phenotypic variability present in the original dataset, which has a reduced amount of images, different preprocessing strategies were employed to enhance model training while preserving clinical relevance and avoiding data leakage. With these strategies it is possible to contrast the influence of preprocessing the original dataset in the performance of the models.

To mitigate the impact of class imbalance, a preprocessing variant utilizes the SMOTEENN (Synthetic Minority Over-sampling Technique combined with Edited Nearest Neighbors) algorithm \cite{manju2019classification}. First SMOTEENN \cite{chawla2002smote} synthetically increases the number of minority class samples by interpolating between existing instances with SMOTE \cite{alejo2010edited}, and then applies data cleaning through Edited Nearest Neighbours (ENN) to remove ambiguous or noisy samples near class boundaries in the training set. This strategy balances the dataset while reducing overfitting risk from overly duplicated data, as observed in other state-of-the-art methods.

\begin{figure}[h!]
    \centering
    \begin{subfigure}{0.48\textwidth}
        \includegraphics[width=\linewidth]{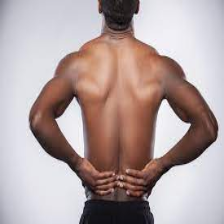}
        \caption{Original}
    \end{subfigure}
    \hfill
    \begin{subfigure}{0.48\textwidth}
        \includegraphics[width=\linewidth, angle=180]{normal8.png}
        \caption{Vertical Flip}
    \end{subfigure}
    \caption{Example of an image from the dataset \cite{bala2023monkeynet} vertical flipping is avoided to preserve anatomical realism, as such transformation would not reflect real-world conditions or would not create new observations of the disease.}
    \label{fig:vertical}
\end{figure}

The other preprocessing approach introduced is data augmentation \cite{chlap2021review}. In this work, data augmentation is applied only to the training set, that would by onward be refered as Data Augmented dataset, in order to increase data diversity without introducing test contamination. Each training image was augmented using random combinations of transformations. Vertical flipping was deliberately excluded to preserve anatomical realism, as certain dermatological conditions and body part asymmetries would render flipped images clinically implausible (Figure \ref{fig:vertical}). Moreover, there are now implementations of CNN capable of extracting features in a rotation-invariant manner \cite{zhang2019rotation}. A total of six new images were generated per original training instance, yielding a substantial increase in training volume. 

\subsection{Model Design}

\begin{figure}[h!]
    \centering
    \includegraphics[width=1\linewidth]{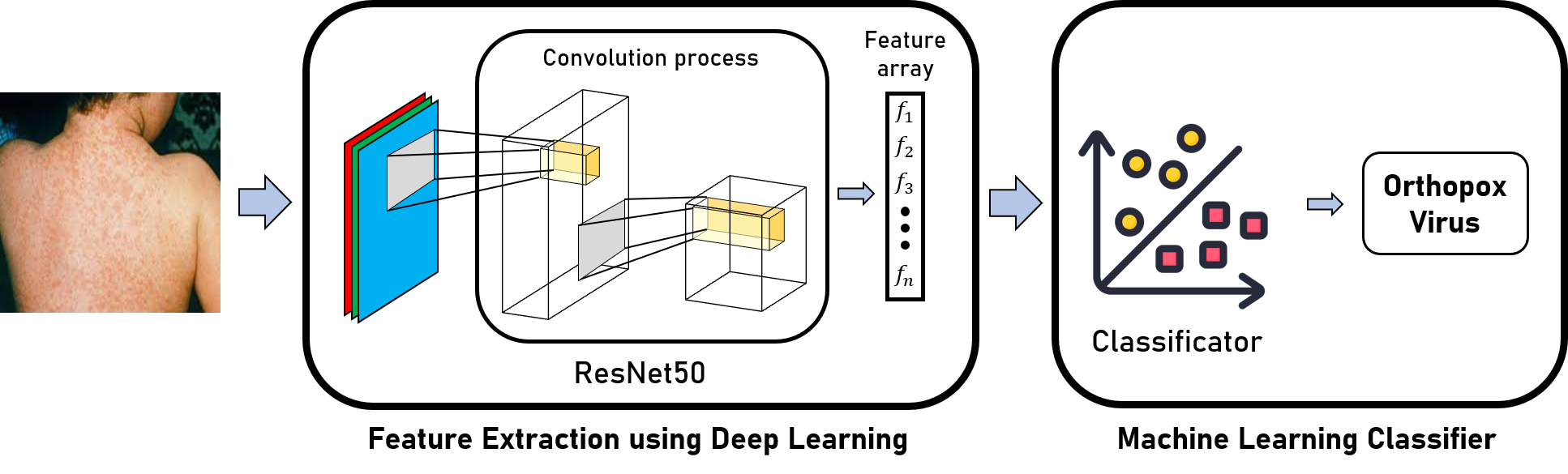}
    \caption{Scheme with the design of the hybrid model for image processing.}
    \label{fig:ensemble}
\end{figure}

A hybrid approach is proposed for the design of the model (Figure \ref{fig:ensemble}). The main idea behind this combination is to have a DL model \cite{lecun2015deep} that does not need to be specifically trained or fine tuned in the problem in order to extract features from the images. With these features, a smaller and less heavy ML model \cite{bishop2006pattern} is trained in the problem instead of also training the DL model. Thus, the training process is accelerated while maintaining the accuracy of the results.

Since the ResNet ANN \cite{he2016deep} is one of the most popular networks in several domains where the input data are images, it is suggested as a DL model. A version of ResNet that has already been trained on the ImageNet general purpose picture dataset \cite{deng2009imagenet} is utilized because the goal is not to train the DL model and optimize the amount of computational resources needed. ResNet is particularly suitable for this task due to its use of residual connections, which allow the training of deep networks without suffering from vanishing gradient problems \cite{borawar2023resnet}. These residual blocks enable neural the model to learn complex hierarchical features that are highly transferable across visual domains. 

Once the deep features are extracted using the pre-trained ResNet 50 model \cite{koonce2021resnet}, traditional ML algorithms are employed to perform the classification task. In this study, Logistic Regression \cite{zhang2020detection}, Multilayer Perceptron (MLP) \cite{valanarasu2022unext}, and Support Vector Machines (SVM) \cite{camlica2015medical} were selected as due to their complementary strengths and proven effectiveness in medical image classification tasks. Logistic Regression offers a simple linear classification approach that performs well when the extracted features are linearly separable, and its interpretability is highly valued in clinical environments. MLP, a type of feedforward artificial neural network, introduces non-linearity and is capable of modeling more complex relationships between features. SVM, on the other hand, is particularly effective in high-dimensional spaces and excels at maximizing the margin between classes, often resulting in strong generalization even with limited data.

\subsection{Experiment Pipeline}

For conducting each experiment, a pipeline has been designed. By following a pipeline for each model and data configuration, results can be generated and compared to know the goodness of the proposed solution.

In the design of this pipeline, the processes followed by the other authors in the state of the art have been taken into account, as well as their limitations. The steps are the following

\begin{enumerate}
    \item Generate an independent and stratified test set to evaluate all conditions on an equal footing and thus reduce the probability of having dependent data between the train and test sets (overestimation of the results). 
    \item Perform each set of experiments training the models with the remaining data: first experiment, with the remaining original data; second experiment, by balancing the data as other authors did; third experiment, by performing Data Augmentation to generate synthetic data.
    \item Finally, for each model, the results are evaluated on the previously extracted final test set. For these evaluations, the relevant statistical tests are performed to better understand the impact of the different combinations on the final results.
\end{enumerate}


\subsection{Model Evaluation}

\begin{figure}[h!]
    \centering
    \includegraphics[width=0.5\linewidth]{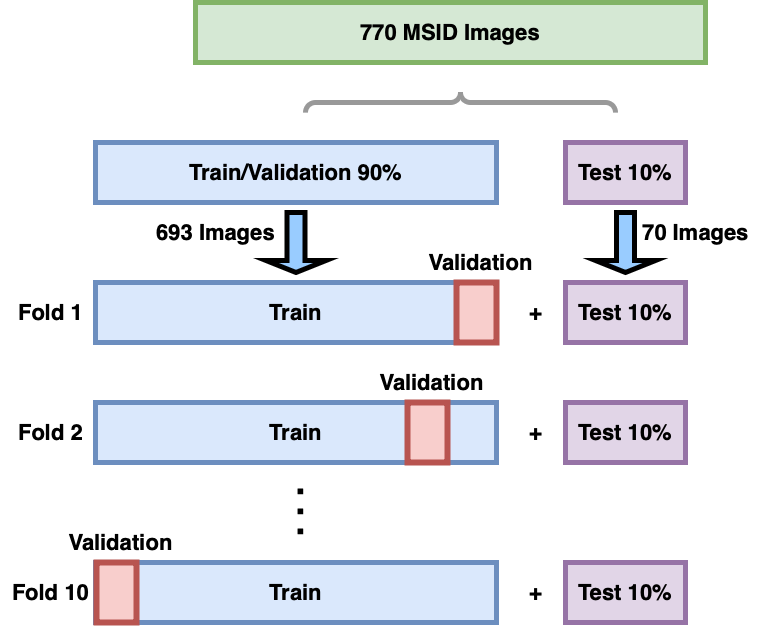}
    \caption{Diagram with the partition of the dataset for training and evaluating each model configuration in each experiment.}
    \label{fig:train}
\end{figure}

For the training and evaluation of each model, a hybrid Stratified Cross-Validation with a Final Hold-Out Test method has been proposed (Figure \ref{fig:train}). Using stratified sampling, the dataset was split into two subsets in order to ensure a reliable assessment of model performance and reduce any potential biases: a training/validation set (the remaining 90\%) and a final test set (10\% of the images) using a Hold-out split strategy \cite{pal2020data}. The test set is fixed and used after fold in the training process in order to evaluate the goodness of the training in new images. The stratified sampling technique was used to maintain the original class distribution inside the test set. All model configurations used in the studies are continuously evaluated using this fixed test set, ensuring an impartial and comparable evaluation of different strategies.

The remaining 90\% of the data (693 pictures) underwent a stratified 10-fold cross-validation \cite{pal2020data}. The dataset was divided into ten equal-sized subsets, or folds, during this process, making sure that each fold retained the class distribution. Nine folds were utilized for training, and the remaining fold was used for validation in each iteration. By avoiding class imbalance problems in the training and validation splits, stratified cross-validation produces more accurate performance estimates.

Before splitting data, all images are  processed through Transfer Learning by a pre-trained DL model to extract new features from the images from the dataset. This automatic feature extraction allows for to reduction of risks in the predictions like bias on feature extraction. The new features are used to train the same model but with different training sets.

Given the size of the dataset (693 photos), using 10 folds is especially appropriate because it balances bias and variance in model evaluation. There would be fewer training samples per iteration with fewer folds, which could result in substantial variance in performance predictions. Conversely, a larger number of folds would increase computational cost without significant benefits in this context. With 10 folds, there is enough data for both training and validation in every iteration, with each training set comprising roughly 623 photos and each validation set comprising roughly 70 images.

Common metrics in AI applied to medical imaging have been used to perform a quantitative evaluation of the results in the test set. These metrics are as follows \cite{puente2020automatic, muller2022towards}: Accuracy (\ref{eq:accuracy}), Recall (\ref{eq:recall}), Precision (\ref{eq:precision}), F1-Score (\ref{eq:f1score}) and Cohen's Kappa (\ref{eq:kappa}).

\begin{equation}
\text{Accuracy} = \frac{TP + TN}{TP + TN + FP + FN}
\label{eq:accuracy}
\end{equation}

\begin{equation}
\text{Recall} = \frac{TP}{TP + FN}
\label{eq:recall}
\end{equation}

\begin{equation}
\text{Precision} = \frac{TP}{TP + FP}
\label{eq:precision}
\end{equation}

\begin{equation}
F_1\text{-Score} = 2 \times \frac{\text{Precision} \times \text{Recall}}{\text{Precision} + \text{Recall}}
\label{eq:f1score}
\end{equation}

\begin{equation}
\kappa = \frac{P_o - P_e}{1 - P_e}
\label{eq:kappa}
\end{equation}

Where:
\begin{itemize}
    \item \( TP \) = True Positives (correctly predicted positive cases)
    \item \( TN \) = True Negatives (correctly predicted negative cases)
    \item \( FP \) = False Positives (incorrectly predicted as positive, also known as Type I error)
    \item \( FN \) = False Negatives (incorrectly predicted as negative, also known as Type II error)
    \item \( P_o \) = Observed accuracy (proportion of correctly classified instances)
    \item \( P_e \) = Expected accuracy (probability of agreement due to chance, based on marginal probabilities)
\end{itemize}

In classification problems involving imbalanced datasets, traditional metrics such as accuracy can be misleading, as they tend to favor the majority class and may not reflect the model’s ability to correctly identify minority classes. To address this limitation, this study emphasizes the use of Cohen’s Kappa as primary evaluation metric. Although, F1-Score provides a balanced measure of a model’s performance in terms of both false positives and false negatives—an essential consideration in medical image analysis where misclassifying less frequent conditions can have serious clinical consequences. Cohen’s Kappa, in turn, accounts for the agreement occurring by chance, offering a more reliable evaluation of model performance, particularly in scenarios with skewed class distributions or imbalanced datasets.

\section{Results}\label{results}

In order to carry out the experiments, a previous experimental design was carried out. With this design, the results can be contrasted with different tested models and with others of the state of the art.

\subsection{Experimental Design}

For the development of the experiments, the proposed pipeline has been followed for each configuration of the evaluated models, using the same test set derived from the original dataset of 770 images. From each image, 100,352 features are extracted, which are less values than the 150,528 values of the pixels of the image. Once the results are obtained, statistical significance tests are conducted to determine which configurations produce statistically significant differences in performance.

Subsequently, the model configuration that demonstrates the best performance is selected, and its results are analyzed using a common test set, after training with resampling and data augmentation techniques, following methodologies employed in other state-of-the-art studies. Finally, these results are also subjected to statistical significance tests to determine whether the proposed model introduces statistically significant improvements compared to other configurations.

A randomized process was followed to select the best parameters for experimentation. This set of parameters can be seen in Table \ref{tab:parameters_summary}.

\begin{table}[h!]
\centering
\caption{Summary of the main parameters used in the experimental configuration.}
\label{tab:parameters_summary}
\begin{tabular}{|l|p{10.5cm}|}
\hline
\textbf{Component} & \textbf{Configuration} \\
\hline

\textbf{Data Augmentation} & 
Vertical flipping with 50\% probability; \newline
Gaussian blur with kernel size 5 applied with 50\% probability; \newline
Vertical flipping was excluded to preserve anatomical consistency. \\
\hline
\textbf{Cross-Validation} & 
Stratified 10-fold cross-validation on 90\% of the dataset; \newline
Independent hold-out test set comprising 10\% of the data. \\
\hline
\textbf{SMOTEENN (Balancing)} & 
Synthetic Minority Over-sampling Technique combined with Edited Nearest Neighbors; \newline
Number of nearest neighbors for oversampling: 5 \\
\hline
\textbf{ResNet} & 
ResNet 50 architecture; \newline
Pretrained with ImageNet; \newline
Removed last layer; \\
\hline
\textbf{Logistic Regression Classifier} & 
Regularization strength (inverse): 0.1; \newline
Class weighting: uniform; \newline
Maximum number of iterations: 326; \newline
Regularization type: L1 (Lasso); \newline
Optimization algorithm: Stochastic Average Gradient Augmented (SAGA). \\
\hline
\textbf{Support Vector Machine (SVM)} & 
Regularization parameter: 100; \newline
Class weighting: uniform; \newline
Kernel function: Sigmoid; \newline
Polynomial degree for kernel: 2; \newline
Kernel coefficient: (1/number of features). \\
\hline
\textbf{Multilayer Perceptron (MLP)} & 
Activation function: logistic sigmoid; \newline
L2 regularization parameter: 0.05; \newline
Hidden layer configuration: one hidden layer with 100 neurons; \newline
Learning rate strategy: inverse scaling; \newline
Optimizer: Adaptive Moment Estimation (Adam). \\
\hline
\end{tabular}
\end{table}

\subsection{Experimental Results}

Experimental results are presented for different dataset configurations. The performance of the models is first analyzed when training with the original dataset consisting of 693 images and tested on a final set of 77 images, followed by an evaluation of the impact of preprocessing techniques on model performance.

\subsubsection{Original Dataset}

Table~\ref{tab:stddev} presents the classification performance of Logistic Regression, SVM, and MLP on the original dataset. Among the evaluated models, Logistic Regression demonstrated the most robust performance across metrics, with the lowest standard deviations, indicating substantial agreement beyond chance and relatively low variability across folds. 

The MLP model achieved the highest mean Kappa, suggesting slightly better average agreement, but also exhibited higher variability ($1.91\%$), which may indicate less consistent generalization across validation folds. Its F1-score was also slightly superior, showing its ability to balance precision and recall effectively, though at the cost of increased variance.

SVM exhibited lower overall performance compared to both Logistic Regression and MLP. While still viable, SVM showed greater variability in key metrics, particularly in precision and F1-score, which suggests less robustness across folds.

\begin{table}[h!]
\centering
\begin{tabular}{c c c c c c}
\hline
\textbf{Model} & \textbf{Accuracy} & \textbf{Precision} & \textbf{Recall} & \textbf{F1-score} & \textbf{Kappa} \\
\hline
\underline{Logistic Reg.} & \underline{90.00\% ± 2.16\%} & \underline{75.17\% ± 12.82\%} & \underline{75.72\% ± 13.39\%} & \underline{75.30\% ± 12.49\%} & \underline{71.19\% ± 1.71\%} \\
MLP & 90.45\% ± 3.16\% & 78.29\% ± 9.65\% & 76.56\% ± 14.93\% & 77.04\% ± 11.56\% & 72.22\% ± 1.91\% \\
SVM & 87.53\% ± 2.74\% & 70.23\% ± 17.21\% & 72.53\% ± 8.09\% & 70.71\% ± 12.38\% & 64.77\% ± 2.92\% \\
\hline
\end{tabular}
\caption{Classification results on Orthopox images using different models. Metrics are reported as mean ± standard deviation, in percentage. The row with the Kappa with the lowest standard deviation value is underlined.}
\label{tab:stddev}
\end{table}

To better assess the reliability of these estimates, Table~\ref{tab:se} presents the same metrics but reported as mean ± standard error (SE), computed from the cross-validation scores of the 10 folds. As expected, the SE values are smaller than the corresponding standard deviations, providing a clearer picture of the precision of the mean estimates. The ranking of models remains consistent, with Logistic Regression again achieving the best results. The reduced uncertainty in the SE values supports the conclusion that Logistic Regression is more stable and robust than the alternatives tested.

\begin{table}[h!]
\centering
\begin{tabular}{c c c c c c}
\hline
\textbf{Model} & \textbf{Accuracy} & \textbf{Precision} & \textbf{Recall} & \textbf{F1-score} & \textbf{Kappa} \\
\hline
\underline{Logistic Reg.} & \underline{90.00\% ± 0.68\%} & \underline{75.17\% ± 4.05\%} & \underline{75.72\% ± 4.23\%} & \underline{75.30\% ± 3.95\%} & \underline{71.19\% ± 0.54\%} \\
MLP & 90.45\% ± 1.00\% & 78.29\% ± 3.05\% & 76.56\% ± 4.72\% & 77.04\% ± 3.66\% & 72.22\% ± 0.60\% \\
SVM & 87.53\% ± 0.87\% & 70.23\% ± 5.44\% & 72.53\% ± 2.56\% & 70.71\% ± 3.92\% & 64.77\% ± 0.92\% \\
\hline
\end{tabular}
\caption{Classification results on Orthopox images using different models. Metrics are reported as mean ± standard error (SE), in percentage, based on 10-fold cross-validation. The row with the Kappa with the lowest error value is underlined.}
\label{tab:se}
\end{table}

\begin{figure}[h!]
    \centering
    \includegraphics[width=0.8\linewidth]{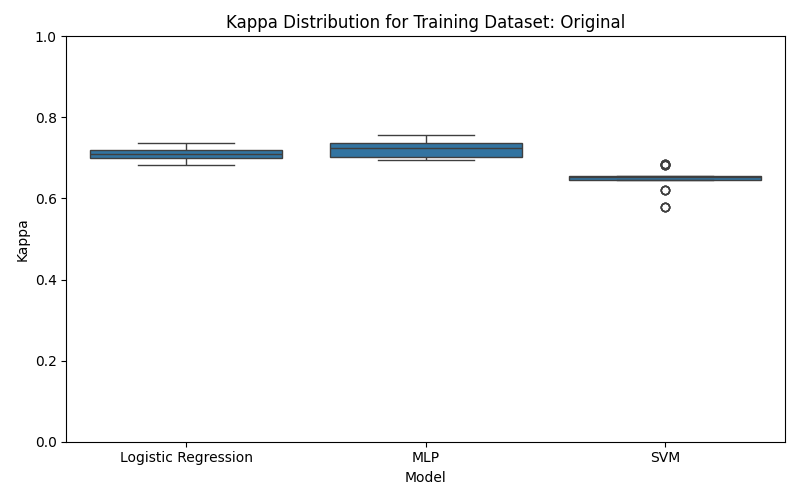}
    \caption{Boxplot of the distributions of the Cohen's Kappa values for each model trained with the original dataset.}
    \label{fig:boxplot_original}
\end{figure}

In order to better understand the goodness of fit of the results based on Cohen's Kappa, a Shapiro-Wilk \cite{hanusz2016shapiro} test was performed in order to find out whether all the distributions of results follow a normal distribution. \textbf{Kappa is preferred as a reference metric because it adjusts for chance agreement, providing a more robust evaluation than raw accuracy, especially in imbalanced classification problems.} As not all of them follow a normal distribution, a Mann-Whitney U significance test \cite{mcknight2010mann} was performed for an $\alpha=5\%$. The only case where there is no statistical significance is between the Logistic Regression and MLP models, where it can be seen that their distributions overlap in a boxplot (Figure \ref{fig:boxplot_original}).

Overall, these results on the original dataset demonstrate that traditional linear classifiers, particularly Logistic Regression, are highly effective in this task when features are extracted with ResNet, even with limited data and no data augmentation. Training these linear classifiers is less resource-consuming than a complete DL model.

\subsubsection{Balanced Dataset}


To further explore the effect of class imbalance on model performance, the training data was resampled to 931 images using the SMOTEENN algorithm in order to achieve a more balanced representation of classes while reducing the risk of overfitting due to synthetic duplication and then it was tested on the same 77 images of the final set. Such balancing strategy is commonly applied in recent state-of-the-art studies. The goal here is to assess whether balancing improves generalization and classifier robustness under conditions similar to those encountered in the state-of-the-art \cite{eliwa2023utilizing}.

When comparing the results in Table~\ref{tab:balanced_stddev} with those from the original dataset in Table~\ref{tab:stddev}, applying SMOTEENN balancing and increasing the training set to 931 images led to more stable and slightly more generalizable models. For instance, Logistic Regression experienced a small decrease in accuracy (from $91.88\% \pm 2.33\%$ to $90.00\% \pm 2.01\%$) but it variability decreased, and it maintained a high F1-score and a consistent Kappa, suggesting improved balance in class-level predictions. MLP also showed competitive performance, slightly surpassing Logistic Regression in some metrics but with higher variability. The SVM classifier, while still the weakest among the three but also improved slightly compared to its performance on the original dataset, particularly in recall and F1-score.

\begin{table}[h!]
\centering
\begin{tabular}{c c c c c c}
\hline
\textbf{Model} & \textbf{Accuracy} & \textbf{Precision} & \textbf{Recall} & \textbf{F1-score} & \textbf{Kappa} \\
\hline
\underline{Logistic Reg.} & \underline{90.00\% ± 2.01\%} & \underline{75.99\% ± 11.48\%} & \underline{76.85\% ± 11.06\%} & \underline{76.03\% ± 9.65\%} & \underline{71.26\% ± 1.58\%} \\
MLP & 89.55\% ± 3.04\% & 76.06\% ± 10.67\% & 73.82\% ± 16.77\% & 74.29\% ± 12.79\% & 69.53\% ± 2.03\% \\
SVM & 87.60\% ± 3.30\% & 71.12\% ± 14.35\% & 72.24\% ± 11.20\% & 71.10\% ± 11.49\% & 64.69\% ± 5.03\% \\
\hline
\end{tabular}
\caption{Classification results on Orthopox images using balanced datasets with SMOTEENN. Metrics are reported as mean ± standard deviation, in percentage. The row with the Kappa with the lowest standard deviation value is underlined.}
\label{tab:balanced_stddev}
\end{table}

A comparison between Table~\ref{tab:balanced_se} and Table~\ref{tab:se} shows that balancing the dataset with SMOTEENN and increasing the training size to 931 images led to slightly improved consistency and reduced uncertainty in classifier performance. Logistic Regression maintained strong results, with nearly unchanged mean accuracy ($90.00\%$ in both cases) and a Kappa of $71.19\% \pm 0.54\%$ on the original data versus $71.26\% \pm 0.50\%$ after balancing—indicating greater stability rather than a shift in central tendency. MLP continued to perform competitively, achieving similar mean values but with slightly higher variability across folds. In contrast, the SVM classifier benefited more clearly from balancing, reducing its error margins and slightly improving its metrics across the board. These outcomes suggest that SMOTEENN not only helps address class imbalance but also contributes to more stable and reliable model estimates.

\begin{table}[h!]
\centering
\begin{tabular}{c c c c c c}
\hline
\textbf{Model} & \textbf{Accuracy} & \textbf{Precision} & \textbf{Recall} & \textbf{F1-score} & \textbf{Kappa} \\
\hline
\underline{Logistic Reg.} & \underline{90.00\% ± 0.64\%} & \underline{75.99\% ± 3.63\%} & \underline{76.85\% ± 3.50\%} & \underline{76.03\% ± 3.05\%} & \underline{71.26\% ± 0.50\%} \\
MLP & 89.55\% ± 0.96\% & 76.06\% ± 3.37\% & 73.82\% ± 5.31\% & 74.29\% ± 4.04\% & 69.53\% ± 0.64\% \\
SVM & 87.60\% ± 1.04\% & 71.12\% ± 4.54\% & 72.24\% ± 3.54\% & 71.10\% ± 3.63\% & 64.69\% ± 1.59\% \\
\hline
\end{tabular}
\caption{Classification results on Orthopox images using balanced datasets with SMOTEEN. Metrics are reported as mean ± standard error (SE), in percentage, based on 10-fold cross-validation. The row with the Kappa with the lowest error value is underlined.}
\label{tab:balanced_se}
\end{table}

\begin{figure}[h!]
    \centering
    \includegraphics[width=0.8\linewidth]{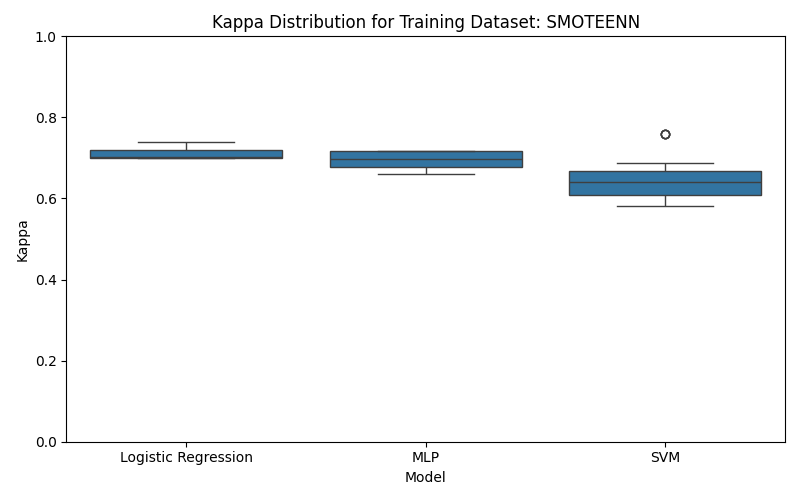}
    \caption{Boxplot of the distributions of the Cohen's Kappa values for each model trained with the dataset balanced with SMOTEENN.}
    \label{fig:boxplot_smoteen}
\end{figure}

Using the same significance test from the previous case, it is found that when balancing the dataset with SMOTEENN, there are already significant differences between them. This may be due to the fact that noise between classes is eliminated. As a result, the classifiers are trained on a cleaner and more representative dataset, which enhances their ability to generalize and allows statistical differences between their performance to emerge more clearly (Figure \ref{fig:boxplot_smoteen}).

Although the application of SMOTEENN increased the training set size from 693 to 931 images (increase of 34\%), the resulting improvements in model performance were relatively modest. While balancing contributed to reducing the variability of most metrics and improving recall in some models, the magnitude of these changes does not appear proportional to the increase in training data size. This suggests that simply increasing the number of examples, does not necessarily translate into substantially better performance.

\subsubsection{Data Augmented Dataset}

To evaluate the impact of increasing the amount of data, synthetic data generation was applied exclusively to the training set in order to avoid information leakage into the validation or test sets. Each original training image was augmented six times (4.158 aditional training images) using a random combination of transformations: vertical flipping with a probability of 0.5, Gaussian blurring with a kernel size of 5 and a probability of 0.5. These transformations were designed to simulate real-world variations without altering the underlying pathology because a considerable amount of images are dependent on their vertical position, so vertical flipping is not used (Figure \ref{fig:vertical}). Although no new clinical features are introduced, this process significantly increases the diversity and volume of the training set, which may help the classifiers generalize better and reduce overfitting. Then, each model is evaluated in the final set of 77 images.

The results presented in Table~\ref{tab:augmented_stddev} show that applying data augmentation led to a noticeable improvement in the performance of both the Logistic Regression and MLP classifiers, while the SVM model remained less competitive. Logistic Regression achieved the lowest standard deviation in Kappa ($\pm 2.12\%$), suggesting a robust and stable performance across cross-validation folds. MLP attained the highest mean values in most metrics, including Kappa ($76.50\%$), indicating strong generalization potential, albeit with slightly higher variability. These improvements are likely due to the increase in training set size and diversity. However, the SVM model continued to perform suboptimally possibly due to sensitivity to high-dimensional feature spaces generated by data augmentation.

\begin{table}[h!]
\centering
\begin{tabular}{c c c c c c}
\hline
\textbf{Model} & \textbf{Accuracy} & \textbf{Precision} & \textbf{Recall} & \textbf{F1-score} & \textbf{Kappa} \\
\hline
\underline{Logistic Reg.} & \underline{91.49\% ± 3.23\%} & \underline{78.70\% ± 15.13\%} & \underline{80.87\% ± 15.07\%} & \underline{79.36\% ± 13.86\%} & \underline{75.68\% ± 2.12\%} \\
MLP & 91.88\% ± 2.85\% & 80.98\% ± 10.02\% & 80.46\% ± 15.65\% & 80.25\% ± 11.57\% & 76.50\% ± 3.11\% \\
SVM & 87.34\% ± 3.50\% & 70.13\% ± 19.56\% & 73.05\% ± 11.80\% & 70.26\% ± 13.51\% & 64.59\% ± 4.84\% \\
\hline
\end{tabular}
\caption{Classification results on Orthopox images using data augmentation. Metrics are reported as mean ± standard deviation, in percentage. The row with the Kappa with the lowest standard deviation value is underlined.}
\label{tab:augmented_stddev}
\end{table}

When analyzing the results in Table~\ref{tab:augmented_se}, Logistic Regression maintains its robust behavior, achieving the lowest SE in Kappa ($\pm 0.67\%$), showing stability across folds. Meanwhile, MLP again delivers the highest mean Kappa ($76.50\%$), but with slightly larger uncertainty ($\pm 0.98\%$), suggesting that while it is the most accurate model on average, it may be more sensitive to fold variability. The SVM classifier continues to lag behind in both performance and consistency, further emphasizing that simpler or less flexible models may not benefit equally from data augmentation techniques.

\begin{table}[h!]
\centering
\begin{tabular}{c c c c c c}
\hline
\textbf{Model} & \textbf{Accuracy} & \textbf{Precision} & \textbf{Recall} & \textbf{F1-score} & \textbf{Kappa} \\
\hline
\underline{Logistic Reg.} & \underline{91.49\% ± 1.02\%} & \underline{78.70\% ± 4.78\%} & \underline{80.87\% ± 4.77\%} & \underline{79.36\% ± 4.38\%} & \underline{75.68\% ± 0.67\%} \\
MLP & 91.88\% ± 0.90\% & 80.98\% ± 3.17\% & 80.46\% ± 4.95\% & 80.25\% ± 3.66\% & 76.50\% ± 0.98\% \\
SVM & 87.34\% ± 1.11\% & 70.13\% ± 6.19\% & 73.05\% ± 3.73\% & 70.26\% ± 4.27\% & 64.59\% ± 1.53\% \\
\hline
\end{tabular}
\caption{Classification results on Orthopox images using data augmentation. Metrics are reported as mean ± standard error (EE), in percentage, based on 10-fold cross-validation. The row with the Kappa with the lowest error value is underlined.}
\label{tab:augmented_se}
\end{table}

\begin{figure}[h!]
    \centering
    \includegraphics[width=0.8\linewidth]{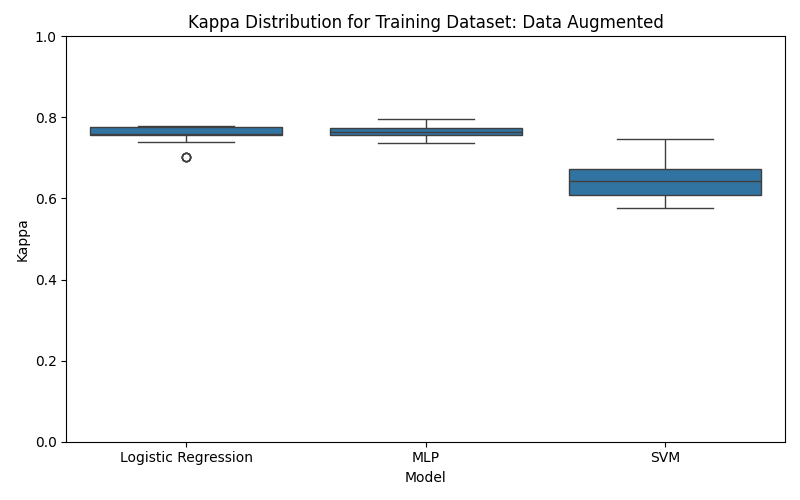}
    \caption{Boxplot of the distributions of Cohen's Kappa values for each model trained with the dataset augmented through image transformations.}
    \label{fig:boxplot_augmented}
\end{figure}

Following the same procedure as in previous experiments, statistically significant differences were found between model performances, except in the case of Logistic Regression versus MLP. In Figure~\ref{fig:boxplot_augmented} the distributions of Logistic Regression and MLP are closely aligned, which is consistent with the lack of statistically significant difference between them, while SVM shows a noticeably lower performance with greater dispersion.

\begin{figure}[h!]
    \centering
    \includegraphics[width=0.9\linewidth]{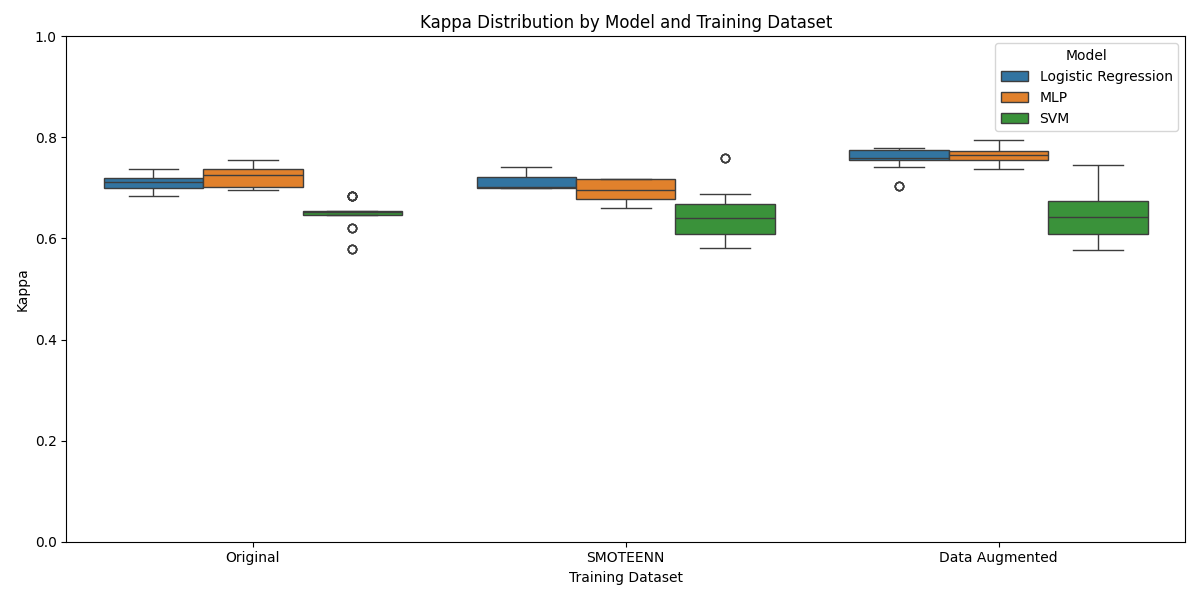}
    \caption{Boxplot of the distributions of Cohen's Kappa of the three experiments.}
    \label{fig:boxplot_total}
\end{figure}

Although the application of data augmentation substantially increased the size of the training dataset, the resulting improvements in model performance were relatively modest. While both Logistic Regression and MLP benefited slightly in terms of average Kappa and reduced variability, the magnitude of the gain does not appear to be proportional to the expansion of the data and the increase on use of computational resources for handling that huge amount of images. This suggests that beyond a certain point, simply adding more synthetic variations of the same images may offer diminishing returns (Figure \ref{fig:boxplot_total}).

Across all experimental conditions, Logistic Regression consistently demonstrates strong and stable performance. This reflects not only its ability to generalize well across folds, but also its robustness in the presence of small datasets and limited variance. Also, it maintains comparable or superior performance while being computationally more efficient and interpretable.

The same Mann-Whitney U tests were applied to the distributions of Cohen's Kappa across the three experimental conditions for the Logistic Regression model. All situations yielded statistically significant differences, indicating that the preprocessing strategy significantly influences the final outcomes.

A comparison with the two best-performing studies on the same dataset is performed: those by Bala et al. \cite{bala2023monkeynet} and Maqsood et al. \cite{maqsood2024mox}. The comparison reveals that their reported results are slightly superior in terms of classification performance. These works have some serious problems with the methodology they used. The test set was extracted after the data augmentation was applied to the whole dataset, as shown in Table~\ref{tab:bala_maqsood_split} according to their publication. In this way, the training set contains information present in the test images, and, consequently, an obvious bias is generated in the test set. This situation results in unrealistic accuracy values as they do not come from an independent evaluation.

\begin{table}[h!]
\centering
\begin{tabular}{c|c|c}
\hline
\textbf{Study} & \textbf{Test Set} & \textbf{Maximum Accuracy} \\
\hline
Bala et al. \cite{bala2023monkeynet} & 1738 images extracted from data augmentation & 97.61\% \\
Maqsood et al. \cite{maqsood2024mox} & 1738 images extracted from data augmentation & 98.64\% \\
\textbf{Propossed method} & \textbf{77 images} extracted  from the \textbf{original dataset} & \textbf{91.49\%} \\
\bottomrule
\end{tabular}
\caption{Comparison of the maximum mean classification accuracy reported in related studies and this work (experiments with data augmentation).}
\label{tab:accuracy_comparison}
\end{table}

\begin{table}[h!]
\centering
\caption{Dataset splitting used in Bala et al. \cite{bala2023monkeynet} and Maqsood et al. \cite{maqsood2024mox} for training, validation, and testing, both with and without data augmentation.}
\label{tab:bala_maqsood_split}
\begin{tabular}{l|ccc|c|ccc|c}

 & \multicolumn{4}{c|}{\textbf{Original Dataset}} & \multicolumn{4}{c}{\textbf{Data Augmented Dataset}} \\
 \textbf{Image Classes} & \textbf{Training} & \textbf{Validation} & \textbf{Testing} & \textbf{Total} & \textbf{Training} & \textbf{Validation} & \textbf{Testing} & \textbf{Total} \\
\hline
Chickenpox & 69 & 17 & 21 & 107 & 1138 & 284 & 355 & 1777 \\
Measles    & 58 & 15 & 18 & 91  & 960  & 241 & 301 & 1502 \\
Monkeypox & 178 & 45 & 56 & 279 & 1691 & 423 & 528 & 2642 \\
Normal     & 187 & 47 & 59 & 293 & 1771 & 443 & 554 & 2768 \\
\hline
\textbf{Total} & 492 & 124 & 154 & 770 & 5560 & 1391 & 1738 & 8689 \\
\hline
\end{tabular}
\end{table}

Furthermore, their final training datasets consist of up to 8,689 images, an order of magnitude larger than the original dataset. The performance improvement compared to baseline models is relatively modest, as in the experiments conducted in this paper. This raises questions about the efficiency and reliability of the data augmentation strategy.

Another concern relates to the types of transformations applied. Both publications report the use of vertical flipping, yet visual analysis has shown that some images in the dataset are anatomically asymmetric or context-dependent (Figure \ref{fig:vertical}), making vertical flipping clinically invalid in certain cases. Applying such transformations indiscriminately may introduce label noise or unrealistic patterns, further undermining the validity of the reported results.

\section{Conclusions and Future Work}\label{conclusions}

This study presents a hybrid Transfer Learning based approach for the classification of orthopoxvirus skin lesions using deep feature extraction from pre-trained convolutional neural networks combined with traditional machine learning models. By leveraging Transfer Learning through ResNet and avoiding the training of deep networks, the proposed method significantly reduces computational cost while maintaining high classification performance across different preprocessing conditions.

An essential aspect of this study is the use of stratified cross-validation combined with a final, fixed test set for evaluation. This strategy ensures that the reported results are not biased by the specific folds used during training and that model performance is assessed on truly unseen data.

Among all evaluated classifiers, Logistic Regression consistently demonstrated strong and stable performance, outperforming or matching more complex models such as MLP and SVM. Its robustness, low variability across folds, and minimal reliance on synthetic data make it an ideal candidate for clinical applications where interpretability, efficiency, and generalization are key.

The experimental results also highlight important considerations regarding data preprocessing. While class balancing (e.g., SMOTEENN) and data augmentation led to slight performance improvements, these gains were not proportional to the increase in data size or complexity. Moreover, comparisons with state-of-the-art methods revealed potential risks of data leakage when augmentation is applied indiscriminately to test sets, which can overestimate real-world performance.

Future work could explore several promising directions like incorporating multimodal information such as patient symptoms, lesion location, or demographic data may help improve classification accuracy by providing contextual information beyond visual features. Another relevant direction involves the use of federated or privacy-preserving learning approaches to train models across distributed data sources without compromising patient confidentiality. 

\section*{Acknowledges}
The authors gratefully acknowledge the Centro de Supercomputación de Galicia (CESGA) for providing access to the Finis Terrae III supercomputer. This work was supported by the Spanish Ministry of Science and Innovation under the projects PID2021-126289OA-I00 and PID2023-149956OB-I00, and by the Xunta de Galicia through project ED431C 2022/46 – Aid for the Consolidation and Structuring of Competitive Research Units (GRC).

\section*{Author Contributions}
\textbf{Alejandro Puente-Castro}: Conceptualization, Methodology, Software, Validation, Formal Analysis, Resources, Data curation, Writing-Original draft preparation, Visualization, Investigation. \textbf{Andres Molares-Ulloa}: Reviewing and Editing. \textbf{Enrique Fernandez-Blanco}: Writing-Reviewing and Editing, Project administration, Funding acquisition. \textbf{Daniel Rivero}: Reviewing and Editing.

\section*{Supplementary Materials}
Source code is available at: \url{https://github.com/TheMVS/Orthopox.git}

\bibliographystyle{plain}
\bibliography{references.bib}  

\begin{thebibliography}{10}

\bibitem{ajmal2023bf2sknet}
Muhammad Ajmal, Muhammad~Attique Khan, Tallha Akram, Abdullah Alqahtani, Majed Alhaisoni, Ammar Armghan, Sara~A Althubiti, and Fayadh Alenezi.
\newblock Bf2sknet: Best deep learning features fusion-assisted framework for multiclass skin lesion classification.
\newblock {\em Neural Computing and Applications}, 35(30):22115--22131, 2023.

\bibitem{alejo2010edited}
Roberto Alejo, Jos{\'e}~Mart{\'\i}nez Sotoca, Rosa~Maria Valdovinos, and P~Toribio.
\newblock Edited nearest neighbor rule for improving neural networks classifications.
\newblock In {\em Advances in Neural Networks-ISNN 2010: 7th International Symposium on Neural Networks, ISNN 2010, Shanghai, China, June 6-9, 2010, Proceedings, Part I 7}, pages 303--310. Springer, 2010.

\bibitem{alharbi2023diagnosis}
Amal~H Alharbi, SK~Towfek, Abdelaziz~A Abdelhamid, Abdelhameed Ibrahim, Marwa~M Eid, Doaa~Sami Khafaga, Nima Khodadadi, Laith Abualigah, and Mohamed Saber.
\newblock Diagnosis of monkeypox disease using transfer learning and binary advanced dipper throated optimization algorithm.
\newblock {\em Biomimetics}, 8(3):313, 2023.

\bibitem{asif2024cgo}
Sohaib Asif, Ming Zhao, Yangfan Li, Fengxiao Tang, and Yusen Zhu.
\newblock Cgo-ensemble: Chaos game optimization algorithm-based fusion of deep neural networks for accurate mpox detection.
\newblock {\em Neural Networks}, 173:106183, 2024.

\bibitem{babkin2022update}
Igor~V Babkin, Irina~N Babkina, and Nina~V Tikunova.
\newblock An update of orthopoxvirus molecular evolution.
\newblock {\em Viruses}, 14(2):388, 2022.

\bibitem{bala2023monkeynet}
Diponkor Bala, Md~Shamim Hossain, Mohammad~Alamgir Hossain, Md~Ibrahim Abdullah, Md~Mizanur Rahman, Balachandran Manavalan, Naijie Gu, Mohammad~S Islam, and Zhangjin Huang.
\newblock Monkeynet: A robust deep convolutional neural network for monkeypox disease detection and classification.
\newblock {\em Neural Networks}, 161:757--775, 2023.

\bibitem{bishop2006pattern}
Christopher~M Bishop and Nasser~M Nasrabadi.
\newblock {\em Pattern recognition and machine learning}, volume~4.
\newblock Springer, 2006.

\bibitem{borawar2023resnet}
Lokesh Borawar and Ravinder Kaur.
\newblock Resnet: Solving vanishing gradient in deep networks.
\newblock In {\em Proceedings of International Conference on Recent Trends in Computing: ICRTC 2022}, pages 235--247. Springer, 2023.

\bibitem{brewka1996artificial}
Gerd Brewka.
\newblock Artificial intelligence—a modern approach by stuart russell and peter norvig, prentice hall. series in artificial intelligence, englewood cliffs, nj.
\newblock {\em The Knowledge Engineering Review}, 11(1):78--79, 1996.

\bibitem{camlica2015medical}
Zehra Camlica, Hamid~R Tizhoosh, and Farzad Khalvati.
\newblock Medical image classification via svm using lbp features from saliency-based folded data.
\newblock In {\em 2015 IEEE 14th international conference on machine learning and applications (ICMLA)}, pages 128--132. IEEE, 2015.

\bibitem{chadaga2023application}
Krishnaraj Chadaga, Srikanth Prabhu, Niranjana Sampathila, Sumith Nireshwalya, Swathi~S Katta, Ru-San Tan, and U~Rajendra Acharya.
\newblock Application of artificial intelligence techniques for monkeypox: a systematic review.
\newblock {\em Diagnostics}, 13(5):824, 2023.

\bibitem{chawla2002smote}
Nitesh~V Chawla, Kevin~W Bowyer, Lawrence~O Hall, and W~Philip Kegelmeyer.
\newblock Smote: synthetic minority over-sampling technique.
\newblock {\em Journal of artificial intelligence research}, 16:321--357, 2002.

\bibitem{chen2025pigmented}
Jinbo Chen, Qian Jiang, Zhuang Ai, Qihao Wei, Sha Xu, Baohai Hao, Yaping Lu, Xuan Huang, and Liuqing Chen.
\newblock Pigmented skin disease classification via deep learning with an attention mechanism.
\newblock {\em Applied Soft Computing}, 170:112571, 2025.

\bibitem{chlap2021review}
Phillip Chlap, Hang Min, Nym Vandenberg, Jason Dowling, Lois Holloway, and Annette Haworth.
\newblock A review of medical image data augmentation techniques for deep learning applications.
\newblock {\em Journal of medical imaging and radiation oncology}, 65(5):545--563, 2021.

\bibitem{das2025novel}
Bihter Das, Huseyin~Alperen Dagdogen, Muhammed~Onur Kaya, and Resul Das.
\newblock A novel hybrid model combining vision transformers and graph convolutional networks for monkeypox disease effective diagnosis.
\newblock {\em Information Fusion}, 117:102858, 2025.

\bibitem{deng2009imagenet}
Jia Deng, Wei Dong, Richard Socher, Li-Jia Li, Kai Li, and Li~Fei-Fei.
\newblock Imagenet: A large-scale hierarchical image database.
\newblock In {\em 2009 IEEE conference on computer vision and pattern recognition}, pages 248--255. Ieee, 2009.

\bibitem{deng2024lsnet}
Xiaodan Deng.
\newblock Lsnet: a deep learning based method for skin lesion classification using limited samples and transfer learning.
\newblock {\em Multimedia Tools and Applications}, pages 1--21, 2024.

\bibitem{diaz2021disease}
James~H Diaz.
\newblock The disease ecology, epidemiology, clinical manifestations, management, prevention, and control of increasing human infections with animal orthopoxviruses.
\newblock {\em Wilderness \& environmental medicine}, 32(4):528--536, 2021.

\bibitem{eliwa2023utilizing}
Entesar Hamed~I Eliwa, Amr~Mohamed El~Koshiry, Tarek Abd El-Hafeez, and Heba~Mamdouh Farghaly.
\newblock Utilizing convolutional neural networks to classify monkeypox skin lesions.
\newblock {\em Scientific reports}, 13(1):14495, 2023.

\bibitem{hanusz2016shapiro}
Zofia Hanusz, Joanna Tarasinska, and Wojciech Zielinski.
\newblock Shapiro--wilk test with known mean.
\newblock {\em REVSTAT-statistical Journal}, 14(1):89--100, 2016.

\bibitem{he2016deep}
Kaiming He, Xiangyu Zhang, Shaoqing Ren, and Jian Sun.
\newblock Deep residual learning for image recognition.
\newblock In {\em Proceedings of the IEEE conference on computer vision and pattern recognition}, pages 770--778, 2016.

\bibitem{hogarty2020artificial}
Daniel~T Hogarty, John~C Su, Kevin Phan, Mohamed Attia, Mohammed Hossny, Saeid Nahavandi, Patricia Lenane, Fergal~J Moloney, and Anousha Yazdabadi.
\newblock Artificial intelligence in dermatology—where we are and the way to the future: a review.
\newblock {\em American journal of clinical dermatology}, 21:41--47, 2020.

\bibitem{koonce2021resnet}
Brett Koonce.
\newblock Resnet 50.
\newblock In {\em Convolutional neural networks with swift for tensorflow: image recognition and dataset categorization}, pages 63--72. Springer, 2021.

\bibitem{kundu2024federated}
Dipanjali Kundu, Md~Mahbubur Rahman, Anichur Rahman, Diganta Das, Umme~Raihan Siddiqi, Md~Golam~Rabiul Alam, Samrat~Kumar Dey, Ghulam Muhammad, and Zulfiqar Ali.
\newblock Federated deep learning for monkeypox disease detection on gan-augmented dataset.
\newblock {\em IEEE Access}, 2024.

\bibitem{lecun2015deep}
Yann LeCun, Yoshua Bengio, and Geoffrey Hinton.
\newblock Deep learning.
\newblock {\em nature}, 521(7553):436--444, 2015.

\bibitem{manju2019classification}
BR~Manju and Anju~R Nair.
\newblock Classification of cardiac arrhythmia of 12 lead ecg using combination of smoteenn, xgboost and machine learning algorithms.
\newblock In {\em 2019 9th International Symposium on Embedded Computing and System Design (ISED)}, pages 1--7. IEEE, 2019.

\bibitem{maqsood2024mox}
Sarmad Maqsood, Robertas Dama{\v{s}}evi{\v{c}}ius, Sana Shahid, and Nils~D Forkert.
\newblock Mox-net: Multi-stage deep hybrid feature fusion and selection framework for monkeypox classification.
\newblock {\em Expert Systems with Applications}, 255:124584, 2024.

\bibitem{mcknight2010mann}
Patrick~E McKnight and Julius Najab.
\newblock Mann-whitney u test.
\newblock {\em The Corsini encyclopedia of psychology}, pages 1--1, 2010.

\bibitem{mitja2023monkeypox}
Oriol Mitj{\`a}, Dimie Ogoina, Boghuma~K Titanji, Cristina Galvan, Jean-Jacques Muyembe, Michael Marks, and Chloe~M Orkin.
\newblock Monkeypox.
\newblock {\em The Lancet}, 401(10370):60--74, 2023.

\bibitem{muller2022towards}
Dominik M{\"u}ller, I{\~n}aki Soto-Rey, and Frank Kramer.
\newblock Towards a guideline for evaluation metrics in medical image segmentation.
\newblock {\em BMC Research Notes}, 15(1):210, 2022.

\bibitem{nuzzo2022declaration}
Jennifer~B Nuzzo, Luciana~L Borio, and Lawrence~O Gostin.
\newblock The who declaration of monkeypox as a global public health emergency.
\newblock {\em Jama}, 328(7):615--617, 2022.

\bibitem{ozdemir2025innovative}
Burhanettin Ozdemir and Ishak Pacal.
\newblock An innovative deep learning framework for skin cancer detection employing convnextv2 and focal self-attention mechanisms.
\newblock {\em Results in Engineering}, 25:103692, 2025.

\bibitem{pal2020data}
Kaushika Pal and Biraj~V Patel.
\newblock Data classification with k-fold cross validation and holdout accuracy estimation methods with 5 different machine learning techniques.
\newblock In {\em 2020 fourth international conference on computing methodologies and communication (ICCMC)}, pages 83--87. IEEE, 2020.

\bibitem{pauli2010orthopox}
Georg Pauli, Johannes Bl{\"u}mel, Reinhard Burger, Christian Drosten, Albrecht Gr{\"o}ner, Lutz G{\"u}rtler, Margarethe Heiden, Martin Hildebrandt, Bernd Jansen, Thomas Montag-Lessing, et~al.
\newblock Orthopox viruses: infections in humans.
\newblock {\em Transfusion Medicine and Hemotherapy}, 37(6):351, 2010.

\bibitem{puente2020automatic}
Alejandro Puente-Castro, Enrique Fernandez-Blanco, Alejandro Pazos, and Cristian~R Munteanu.
\newblock Automatic assessment of alzheimer’s disease diagnosis based on deep learning techniques.
\newblock {\em Computers in biology and medicine}, 120:103764, 2020.

\bibitem{rahmani2024monkeypox}
Erfan Rahmani, Ziba Bayat, Mehrdad Farrokhi, Shiva Karimian, Reza Zahedpasha, Hamed Sabzehie, Sepehr~Ramezani Poor, Parisa~Jafari Khouzani, Solmaz Aminpour, Mohammad Karami, et~al.
\newblock Monkeypox: A comprehensive review of virology, epidemiology, transmission, diagnosis, prevention, treatment, and artificial intelligence applications.
\newblock {\em Archives of Academic Emergency Medicine}, 12(1):e70, 2024.

\bibitem{sarwar2024skin}
Nadeem Sarwar, Asma Irshad, Qamar~H Naith, Kholod D.~Alsufiani, and Faris~A Almalki.
\newblock Skin lesion segmentation using deep learning algorithm with ant colony optimization.
\newblock {\em BMC Medical Informatics and Decision Making}, 24(1):265, 2024.

\bibitem{shchelkunov2013increasing}
Sergei~N Shchelkunov.
\newblock An increasing danger of zoonotic orthopoxvirus infections.
\newblock {\em PLoS pathogens}, 9(12):e1003756, 2013.

\bibitem{soe2024evaluation}
Nyi~N Soe, Zhen Yu, Phyu~M Latt, David Lee, Jason~J Ong, Zongyuan Ge, Christopher~K Fairley, and Lei Zhang.
\newblock Evaluation of artificial intelligence-powered screening for sexually transmitted infections-related skin lesions using clinical images and metadata.
\newblock {\em BMC medicine}, 22(1):296, 2024.

\bibitem{tahir2023dscc_net}
Maryam Tahir, Ahmad Naeem, Hassaan Malik, Jawad Tanveer, Rizwan~Ali Naqvi, and Seung-Won Lee.
\newblock Dscc\_net: multi-classification deep learning models for diagnosing of skin cancer using dermoscopic images.
\newblock {\em Cancers}, 15(7):2179, 2023.

\bibitem{tembhurne2023skin}
Jitendra~V Tembhurne, Nachiketa Hebbar, Hemprasad~Y Patil, and Tausif Diwan.
\newblock Skin cancer detection using ensemble of machine learning and deep learning techniques.
\newblock {\em Multimedia Tools and Applications}, 82(18):27501--27524, 2023.

\bibitem{valanarasu2022unext}
Jeya Maria~Jose Valanarasu and Vishal~M Patel.
\newblock Unext: Mlp-based rapid medical image segmentation network.
\newblock In {\em International conference on medical image computing and computer-assisted intervention}, pages 23--33. Springer, 2022.

\bibitem{vuran2025multi}
Seyfettin Vuran, Murat Ucan, Mehmet Akin, and Mehmet Kaya.
\newblock Multi-classification of skin lesion images including mpox disease using transformer-based deep learning architectures.
\newblock {\em Diagnostics}, 15(3):374, 2025.

\bibitem{zha2025data}
Daochen Zha, Zaid~Pervaiz Bhat, Kwei-Herng Lai, Fan Yang, Zhimeng Jiang, Shaochen Zhong, and Xia Hu.
\newblock Data-centric artificial intelligence: A survey.
\newblock {\em ACM Computing Surveys}, 57(5):1--42, 2025.

\bibitem{zhang2020detection}
Zheng Zhang and Yibo Han.
\newblock Detection of ovarian tumors in obstetric ultrasound imaging using logistic regression classifier with an advanced machine learning approach.
\newblock {\em IEEE Access}, 8:44999--45008, 2020.

\bibitem{zhang2019rotation}
Zhiyuan Zhang, Binh-Son Hua, David~W Rosen, and Sai-Kit Yeung.
\newblock Rotation invariant convolutions for 3d point clouds deep learning.
\newblock In {\em 2019 International conference on 3d vision (3DV)}, pages 204--213. IEEE, 2019.

\end{thebibliography}

\end{document}